\documentclass[10pt,twocolumn,letterpaper]{article}

\usepackage{cvpr}              

\usepackage{graphicx}
\usepackage{amsmath}
\usepackage{amssymb}
\usepackage{booktabs}
\usepackage{multirow}
\usepackage{xcolor}

\usepackage[pagebackref,breaklinks,colorlinks]{hyperref}

\usepackage[capitalize]{cleveref}
\crefname{section}{Sec.}{Secs.}
\Crefname{section}{Section}{Sections}
\Crefname{table}{Table}{Tables}
\crefname{table}{Tab.}{Tabs.}

\def\cvprPaperID{*****} 
\def\confName{CVPR}
\def\confYear{2022}

\newcommand{\gray}{\textcolor{gray}}
\newcommand{\ANURAG}[1]{\textcolor{blue}{\bf \small [Anurag: #1]}}
\newcommand{\XXMAN}[1]{\textcolor{red}{\bf \small [Xuehan: #1]}}
\newcommand{\ARSHA}[1]{\textcolor{green}{\bf \small [Arsha: #1]}}
\newcommand{\CORDELIA}[1]{\textcolor{magenta}{\bf \small [Cordelia: #1]}}

\renewcommand{\ANURAG}[1]{\textcolor{magenta}{{}}}
\renewcommand{\XXMAN}[1]{\textcolor{magenta}{{}}}
\renewcommand{\ARSHA}[1]{\textcolor{magenta}{{}}}
\renewcommand{\CORDELIA}[1]{\textcolor{magenta}{{}}}

\begin{document}

\title{M\&M Mix: A Multimodal Multiview Transformer Ensemble}

\author{Xuehan Xiong, Anurag Arnab, Arsha Nagrani, Cordelia Schmid\\
Google Research\\
{\tt\small \{xxman,aarnab,anagrani,cordelias\}@google.com}
}
\maketitle

\begin{abstract}
   This report describes the approach behind our winning solution to the 2022 Epic-Kitchens Action Recognition Challenge\ARSHA{Do we need to put the team name in the abstract? Just because Xuehan you and I aren't in Grenoble ...}\XXMAN{We need to mention the team name said in the challenge.}. Our approach builds upon our recent work, Multiview Transformer for Video Recognition (MTV), and adapts it to multimodal inputs. Our final submission consists of an ensemble of Multimodal MTV (M\&M) models varying backbone sizes and input modalities. Our approach achieved 52.8\% Top-1 accuracy on the test set in action classes, which is 4.1\% higher than last year’s winning entry.
   \ANURAG{Can we say that we are the winning entry?}\XXMAN{No, it is not official yet.}
\end{abstract}

\section{Introduction}
\label{sec:intro}
Transformers have replaced Convolutional Networks (CNNs) as the de facto backbone for video understanding. State-of-the-art results on popular datasets (e.g., Kinetics~\cite{Carreira_2017_CVPR}, Moments in Time~\cite{monfort_pami_2019}, Epic-Kitchens~\cite{damen2021rescaling}, etc) are all obtained using a pure transformer-based approach. Our approach is built upon a very recent state-of-the-art method for video classification, Multiview Transformers for Video Recognition (MTV)~\cite{yan2022multiview}. MTV proposed a multi-stream architecture to process video data in a multiscale fashion where each stream takes in different-sized tubelets of RGB frames, however no other modalities (such as sound) were used for making a prediction. 

Epic-Kitchens is a large-scale dataset of first-person (egocentric) videos recorded in kitchen environments. Contestants of the Action Recognition challenge are required to predict a verb and a noun for each video clip. Videos in this dataset are multimodal (they contain an audio track) and the egocentric domain consists of rich sounds resulting from the interactions between humans and objects, as well as the proximity of the wearable microphone to the undergoing action. Sound is a hence a discriminative feature for identifying actions~\cite{kazakos2019epic, nagrani2021attention}, for example, the sound of running water provides important cues to predict actions such as ``wash glass’’. Optical flow is another modality that is complementary to RGB frames as shown in previous work~\cite{simonyan_neurips_2014}. As we will show later in the experiments, this observation remains true for state-of-the-art video transformer models,  such as MTV. In this work, we extend MTV to process multimodal inputs where each stream encodes input data from one temporal resolution and from one modality.

\section{Multimodal Multiview Transformers}
\label{sec:method}
\subsection{Background (MTV)}
\begin{figure}[t]
    \vspace{-\baselineskip}
	\centering
    \includegraphics[width=0.9\columnwidth]{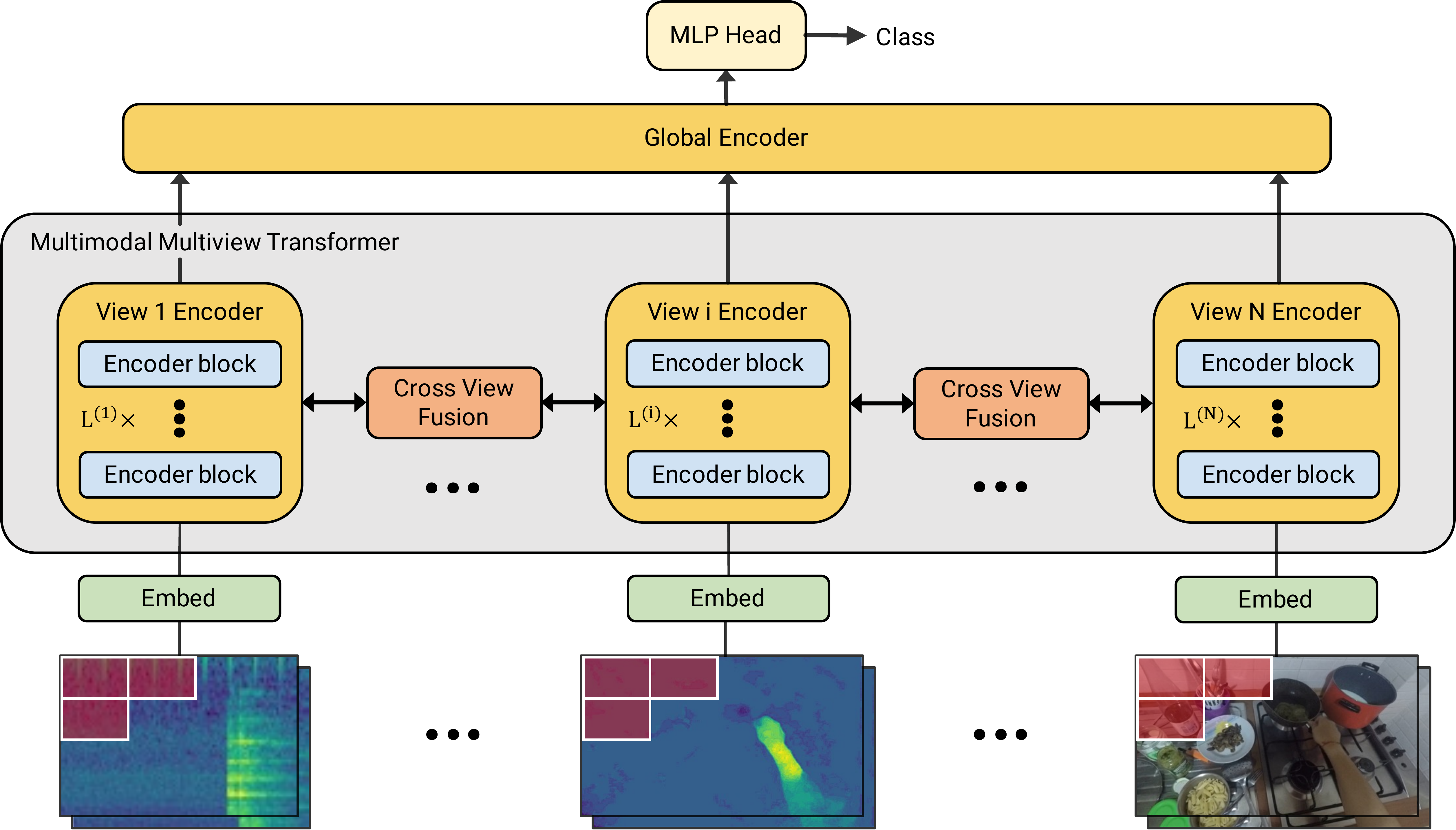}
    \caption{
    Overview of our Multimodal Multiview Transformer (M\&M). The input video consists of three modalities, spectrogram, optical flow, and RGB frames (from left to right) and we create multiple representations or ``views’’ by tokenizing each input modality using tubelets of different sizes. These tokens are fed into separate encoders and further fused through a Cross View Fusion module, and finally aggregated by a global encoder. Note that each encoder can vary in architecture.
    }
    \vspace{-\baselineskip}
	\label{fig:model-overview}
\end{figure}
This section presents a brief overview of Multiview Transformers (MTV)~\cite{yan2022multiview}. It consists of separate transformer encoders for each view which are connected by lateral connections to fuse cross-view information.
A view is defined as a video representation expressed by a set of fixed-sized tubelets. A larger view corresponds to a set of larger tubelets (and thus fewer tokens) and a smaller view corresponds to smaller tubelets (and thus more tokens).
Each transformer layer within the encoders follows the same design as the original transformer of Vaswani et al.~\cite{vaswani2017attention}. Furthermore, within each transformer layer, self-attention is computed only among tokens extracted from the same temporal index, following the Factorised Encoder of~\cite{arnab2021vivit}. This significantly reduces the computational cost of the model. We chose cross-view attention as the fusion method as it gives the best performance as shown in~\cite{yan2022multiview}.
Finally, the classification tokens from each view are extracted and processed with another transformer encoder that aggregates information from all views. 

\subsection{M\&M}
The overall architecture of M\&M (shown in Figure.~\ref{fig:model-overview}) remains the same as MTV except for the input tokenization step. In this example, the input video has three modalities, RGB, optical flow, and short-term magnitude spectrograms derived from audio. For each modality, we can have multiple representations or ``views’’ by tokenizing the frames from this modality using different tubelet sizes. An alternative design is to use a single encoder that takes in tokens from all modalities~\cite{gabeur2020multi, kazakos2021little, nagrani2021attention}. Our design of utilizing a separate encoder for each multimodal view is more flexible. As Yan et al.~\cite{yan2022multiview} have shown, it is sufficient to use a smaller encoder to learn representations from larger views of RGB frames. Feichtenhofer et al.~\cite{feichtenhofer_iccv_2019} applied a smaller CNN (e.g., a smaller number of channels) to learn motion information and a larger one for encoding the semantics of frames.
One advantage of our design is that our architecture also supports multiscale processing within each modality.

\section{Experiments}
\subsection{Experimental setup}
\paragraph{Model notation} For the backbone of each view, we consider four ViT variants, ``Tiny'', ``Small'', ``Base'', and ``Large''. Their settings strictly follow the ones defined in BERT~\cite{devlin_naacl_2019} and ViT~\cite{dosovitskiy2020image}, \ie number of transformer layers, number of attention heads, hidden dimensions.
For convenience, each model variant is denoted with the following abbreviations indicating the backbone size, tubelet length, and input modality.
For example, B/2:R+S/4:S+Ti/8:F denotes a three-view model, where a ``Base'', ``Small'', and ``Tiny'' encoders are used to processes tokens from RGB tubelets of sizes $16\times16\times2$, spectrogram tubelets of sizes $16\times16\times4$, and optical flow tubelets of sizes $16\times16\times8$, respectively.
Note that we omit 16 in our model abbreviations because all our models use $16\times16$ as the spatial tubelet size following ViT~\cite{dosovitskiy2020image}. If we omit the modality in the notation, we assume all views use RGB frames as the modality.
All model variants use the same global encoder which follows the ``Base'' architecture, except that the number of heads is set to 8 instead of 12. The reason is that the hidden dimension of the tokens should be divisible by the number of heads for multi-head attention, and the number of hidden dimensions across all standard transformer architectures (from ``Tiny'' to ``Large''~\cite{steiner2021augreg, dosovitskiy2020image}) is divisible by 8.

\paragraph{Optical flow and spectrogram extraction} 
We compute optical flow using the FlowNet~\cite{dosovitskiy2015flownet} algorithm. Audio spectrograms are extracted in a similar manner to~\cite{hershey2017cnn}. All audio is converted to monochannel and resampled to 16kHz. Spectrograms are then extracted using short-term Fourier transforms with a Hann window of 25ms with 15ms hop. The resulting spectrogram is integrated into
64 mel-spaced frequency bins (lower cutoff 125 Hz and upper corner frequency 7500 Hz) and the squared magnitude is extracted. This gives us mel spectrograms of 96 $\times$ 64 bins for 0.96 seconds of audio. For the entire clip, we run the above procedure in a sliding window fashion with a temporal hop of 40ms to align with RGB frame rate (25FPS). Spectrograms are normalized to [-1, 1] before feeding into the model.

\paragraph{Initialization} We trained two RGB-only models B/2+S/4+Ti/8 and L/2+B/4+S/8+Ti/16 on WTS~\cite{stroud2020learning} and use them to initialize multimodal models. Optical flow images have two input channels and spectrogram images only have one so the initial tubelet embedding layer has a different shape than the pretrained RGB models.
To address this issue, we simply average the kernel of the embedding layer along the input channel axis and perform tiling.

\paragraph{Training and inference}
\begin{table} 
	\centering
	\scriptsize{
\begin{tabular}{  l  c }
    \toprule
    \multicolumn{2}{l}{\textit{Data augmentation}}  \\
    Random crop probability & 1.0 \\
    Random flip probability & 0.5\\
    Scale jitter probability	& 1.0 \\
    Maximum scale			  & 1.33 \\
    Minimum scale			  & 0.9 \\
    Colour jitter probability  & 0.8 \\
    Rand augment number of layers~\cite{cubuk_arxiv_2019}		 & 3 \\
    Rand augment magnitude~\cite{cubuk_arxiv_2019} & 10  \\
    \midrule
    \multicolumn{2}{l}{\textit{Regularisation}} \\
    Stochastic droplayer rate~\cite{huang_stochasticdepth_eccv_2016} & 0.1 \\
    Label smoothing~\cite{szegedy_cvpr_2016} & 0.1 \\
    \bottomrule
   \end{tabular}
   }
	\caption{Data augmentation and regularization parameters.}
\label{tab:hparams}
\end{table}

All models are trained on 64 frames with a temporal stride of 1. In Epic-Kitchens, each video is labeled with a ``verb'' and a ``noun''. We predict both categories using a single network with two ``heads''.
We train all our models for 50 epochs with a global batch size of 128 using synchronous SGD with momentum of 0.9 following a cosine learning rate schedule with a linear warm up. The initial learning rates for all models are set to 0.4. We follow~\cite{arnab2021vivit,yan2022multiview,dehghani2021scenic} and apply the same data augmentation and regularization schemes~\cite{huang_stochasticdepth_eccv_2016,cubuk_arxiv_2019,szegedy_cvpr_2016}, which were used by~\cite{touvron2021training} to train vision transformers more effectively. For spectrograms we use SpecAugment~\cite{park2019specaugment} with a max time mask length of 96 frames and max frequency mask length of 16 bins following MBT~\cite{nagrani2021attention}. \ARSHA{Xuehan check these hyperparams}\XXMAN{Changed 48 to 16 and 192 to 96.} See Table~\ref{tab:hparams} for detailed settings.
During single-model inference, we adopt the standard evaluation protocol by averaging over four temporal crops. To produce the final predictions from the model ensemble, we simply average the logits produced by each model.

\subsection{Ablation study} 
\label{sec:ablations}
We use a RGB-only model B/2+S/4+Ti/8 for the studies in Table~\ref{tab:pretraining} and ~\ref{tab:resolution}. We report Top-1 accuracies on Action, Noun, and Verb classes obtained from averaging predictions across four temporal crops. All numbers reported in this section are from the validation set.

\paragraph{Effects of pretraining}
Table~\ref{tab:pretraining} presents the finetuning results from models pretrained on Kinetics 400~\cite{kay_arxiv_2017}, Kinetics 700~\cite{kay_arxiv_2017}, and WTS~\cite{stroud2020learning} datasets. Kinetics 400 and 700 consist of 230,000 and 530,000 10s video clips focusing on human actions with each clip labeled with one of the 400 and 700 classes, respectively. WTS contains 60M videos with only video-level labels. All three pretraining datasets are from a different domain than Epic-Kitchens that is composed of egocentric videos. Table~\ref{tab:pretraining} shows that it is more beneficial to pretrain on a large-scale weakly supervised dataset than on a smaller set of trimmed video clips.
\begin{table} 
	\centering
	\scriptsize{
\begin{tabular}{  c  c  c  c }
    \toprule
    Pretraining datasets & Top-1 Action & Top-1 Noun & Top-1 Verb \\
    \midrule
    K400 & 46.7 & 60.5 & 67.8 \\
    K700 & 48.0 & 61.2 & 69.1 \\
    WTS & 49.3 & 63.0 & 69.4 \\
    \bottomrule
   \end{tabular}
   }
	\caption{Effects of different pretraining datasets. All models are trained and evaluated on $224\times224$ crops.}
\label{tab:pretraining}
\end{table}

\paragraph{Effects of input resolution}
As Table~\ref{tab:resolution} shown, as spatial resolution increases so does top-1 accuracy for nouns. Accuracies for verbs are also improved and this is likely due to the increased number of tokens that help the model better understand motion in the scene.

\begin{table} 
	\centering
	\scriptsize{
\begin{tabular}{  c  c  c  c }
    \toprule
    Spatial resolution & Top-1 Action & Top-1 Noun & Top-1 Verb \\
    \midrule
    224p & 49.3 & 63.0 & 69.4 \\
    280p & 50.5 & 63.9 & 69.9 \\
    432p & 52.7 & 66.1 & 71.2 \\
    \bottomrule
   \end{tabular}
   }
	\caption{Effects of increasing spatial resolution. All models are finetuned from a WTS-pretrained checkpoint.}
\label{tab:resolution}
\end{table}

\paragraph{Effects of combining different modalities}
\begin{table} 
	\centering
	\scriptsize{
\begin{tabular}{  l  c  c  c }
    \toprule
    Models & Top-1 Action & Top-1 Noun & Top-1 Verb \\
    \midrule
    B/2:R+S/4:R+Ti/8:R & 52.7 & 66.1 & 71.2 \\
    B/2:F+S/4:F+Ti/8:F & 40.5 & 50.1 & 68.1 \\
    \midrule
    B/2:R+S/4:F+Ti/8:R & 53.4 & \bf{66.5} & 71.9 \\
    B/2:R+S/4:S+Ti/8:R & 53.2 & 66.3 & \bf{72.0} \\
    \midrule
    B/2:R+S/4:S+Ti/8:F & \bf{53.6} & 66.3 & \bf{72.0} \\
    \bottomrule
   \end{tabular}
   }
	\caption{Effects of combining different modalities. All models are trained and evaluated on $432\times432$ crops.
	As an example of our naming convention, B/2:R+S/4:S+Ti/8:F denotes a three-view model, where a ``Base'', ``Small'', and ``Tiny'' encoders are used to processes tokens from RGB tubelets of sizes $16\times16\times2$, spectrogram tubelets of sizes $16\times16\times4$, and optical flow tubelets of sizes $16\times16\times8$, respectively.
}
\label{tab:modalities}
\end{table}

The first two rows in Table~\ref{tab:modalities} present the Top-1 accuracies of the RGB-only and the Flow-only models. Changing input modality of the ``Small'' encoder from RGB to flow and to spectrogram improves Top-1 accuracy on action from 52.7 to 53.4 and 53.2, respectively. Combining all three modalities gives the best performance on action with a score of 53.6. All models share similar FLOPs with the only difference being the initial embedding layers. RGB is the most informative modality for predicting ``nouns'', there is little gain by adding flow and audio. 
However, optical flow and audio provide complimentary information to RGB for predicting ``verbs''.
 
\subsection{Comparison to the state-of-the-art}
\label{sec:sota}
\begin{table} 
	\centering
	\scriptsize{
\begin{tabular}{c  l  c  c  c }
    \toprule
    Data split & Models & Top-1 Action & Top-1 Noun & Top-1 Verb \\
    \midrule
    \multirow{4}{*}{validation} & MoViNet~\cite{kondratyuk2021movinets} & 47.7 & 57.3 & \bf{72.2} \\
    & MeMViT~\cite{wu2022memvit} & 48.4 & 60.3 & 71.4 \\
    & Omnivore~\cite{girdhar2022omnivore} & 49.9 & 61.7 & 69.5 \\
    & \bf{M\&M-B} & \bf{53.6} & \bf{66.3} & 72.0 \\
    \midrule
    \multirow{2}{*}{test} & \gray{2021 winner \cite{Damen2021CHALLENGES}} & \gray{48.7} & \gray{59.2} & \gray{\bf{70.6}} \\
    & \bf{M\&M-B} & \bf{49.6} & \bf{63.7} & 68.0 \\
    \bottomrule
   \end{tabular}
   }
	\caption{Comparisons to state-of-the-art. M\&M-B refers to our three-view multimodal MTV model, B/2:R+S/4:S+Ti/8:F (no ensembling). The gray row is the winning entry from last year's challenge, which uses a 6-model ensemble. All other rows are from a single-model evaluation.}
\label{tab:sota}
\end{table}

Table~\ref{tab:sota} compares our best single model to the previous state-of-the-art on the Epic-Kitchens dataset and last year's winning entry of the challenge. Our M\&M-B model improves over the previous state-of-the-art~\cite{girdhar2022omnivore} by a margin of 3.7\% in Top-1 action accuracy and also outperforms last year's winning method~\cite{Damen2021CHALLENGES}, which uses a 6-model ensemble.

\subsection{Model ensemble}
\label{sec:ensemble}
\begin{table*} 
	\centering
	\scriptsize{
\begin{tabular}{  l l  l c c  c c }
    \toprule
    Model indices & Model variants & Pretraining datasets & Resolution & Top-1 Action & Top-1 Noun & Top-1 Verb \\
    \midrule
    0 & B/2:R+S/4:R+Ti/8:F & WTS $\to$ K700 & 432p & 53.4 & 66.4 & 71.8 \\
    1 & B/2:R+S/4:F+Ti/8:R & WTS $\to$ K700 & 432p & 53.4 & 66.5 & 71.9 \\
    2 & L/2:R+B/4:F+S/8:F+Ti/16:R & WTS $\to$ K700 & 320p & 53.0 & 66.7 & 71.1 \\
    3 & L/2:R+B/4:R+S/8:R+Ti/16:R & WTS & 352p & 52.6 & 67.2 & 69.8 \\
    4 & B/2:F+S/4:F+Ti/8:F & WTS $\to$ K700 & 432p & 40.5 & 50.1 & 68.1 \\
    5 & B/2:R+S/4:R+Ti/8:R (128$\times$1) & WTS & 304p & 52.4 & 65.6 & 71.3 \\
    6 & L/2:F+B/4:F+S/8:F+Ti/16:F & WTS $\to$ K700 & 352p & 40.9 & 50.6 & 67.2 \\
    7 & L/2:R+B/4:F+S/8:S+Ti/16:R & WTS & 320p & 53.6 & 67.0 & 71.7 \\
    8 & B/2:R+S/4:S+Ti/8:F & WTS & 432p & 53.6 & 66.3 & 72.0 \\
    9 & B/2:R+S/4:S+Ti/8:R & WTS & 432p & 53.2 & 66.3 & 72.0 \\
    10 & B/2:R+S/4:R+Ti/8:S & WTS & 432p & 53.4 & 66.6 & 72.0 \\
    \bottomrule
   \end{tabular}
   }
	\caption{All model variants used in our final ensemble and their respective performance on the validation set. WTS$\to$K700 denotes a pretraining strategy where we first pretrain the model on WTS and then finetune on Kinetics 700. Model 5 is trained and evaluated on 128 frames instead of 64 for all other models.}
\label{tab:all_models}
\end{table*}

\begin{table} 
	\centering
	\scriptsize{
\begin{tabular}{  l  c  c  c }
    \toprule
    Model indices & Top-1 Action (val/test) & Top-1 Noun & Top-1 Verb \\
    \midrule
    0,1,2,3,5,6,7,8,9,10 & \multirow{2}{*}{56.9/52.8} & 69.2/66.2 &  \\
    4,5,6,7,8,9,10 & &  & 75.0/70.9 \\
    \bottomrule
   \end{tabular}
   }
	\caption{Results from our final model ensemble on both validation/test sets. Different sets of models are used for predicting nouns and verbs.}
\label{tab:ensemble}
\end{table}

To create the final submission, we generated two model ensembles one for predicting the verbs and the other for nouns. Table~\ref{tab:all_models} lists all individual models used in this challenge and their corresponding performance on the validation set. Table~\ref{tab:ensemble} shows which models we used for verbs and nouns. Using this model ensembling strategy, we improve the Top-1 action accuracy from 53.6 (from our single best model) to 56.9 on the validation set. Our final submission scored 52.8 on Epic-Kitchens test set, which is 4.1\% higher than last year's winning entry.

\ANURAG{Would our single model have beaten last year's entry, or this year's second place?}\XXMAN{Yes, I mentioned that in the SOTA section.}

\section{Conclusions}
In this report, we present the approach behind our submission to the 2022 Epic-Kitchens Action Recognition challenge. We proposed M\&M, a transformer backbone that learns a multimodal and multiscale representation of videos. Our final submission is an ensemble of M\&M models with varying backbone sizes and modality mixes. It scored 52.8 in top-1 accuracy on action classes on the test set, which is 4.1\% higher than the last year’s winner.

{\small
\bibliographystyle{ieee_fullname}
\bibliography{main}

\begin{thebibliography}{10}\itemsep=-1pt

\bibitem{arnab2021vivit}
Anurag Arnab, Mostafa Dehghani, Georg Heigold, Chen Sun, Mario Lučić, and
  Cordelia Schmid.
\newblock Vivit: A video vision transformer.
\newblock In {\em ICCV}, 2021.

\bibitem{Carreira_2017_CVPR}
Joao Carreira and Andrew Zisserman.
\newblock Quo vadis, action recognition? a new model and the kinetics dataset.
\newblock In {\em CVPR}, 2017.

\bibitem{cubuk_arxiv_2019}
Ekin~D. Cubuk, Barret Zoph, Jonathon Shlens, and Quoc~V. Le.
\newblock Randaugment: Practical automated data augmentation with a reduced
  search space.
\newblock In {\em NeurIPS}, 2020.

\bibitem{damen2021rescaling}
Dima Damen, Hazel Doughty, Giovanni~Maria Farinella, , Antonino Furnari, Jian
  Ma, Evangelos Kazakos, Davide Moltisanti, Jonathan Munro, Toby Perrett, Will
  Price, and Michael Wray.
\newblock Rescaling egocentric vision: Collection, pipeline and challenges for
  epic-kitchens-100.
\newblock In {\em IJCV}, 2021.

\bibitem{Damen2021CHALLENGES}
Dima Damen, Adriano Fragomeni, Jonathan Munro, Toby Perrett, Daniel Whettam,
  Michael Wray, Antonino Furnari, Giovanni~Maria Farinella, and Davide
  Moltisanti.
\newblock Epic-kitchens-100- 2021 challenges report.
\newblock Technical report, University of Bristol, 2021.

\bibitem{dehghani2021scenic}
Mostafa Dehghani, Alexey Gritsenko, Anurag Arnab, Matthias Minderer, and Yi
  Tay.
\newblock {Scenic}: A {JAX} library for computer vision research and beyond.
\newblock In {\em arXiv preprint arXiv:2110.11403}, 2021.

\bibitem{devlin_naacl_2019}
Jacob Devlin, Ming-Wei Chang, Kenton Lee, and Kristina Toutanova.
\newblock {BERT}: Pre-training of deep bidirectional transformers for language
  understanding.
\newblock In {\em NAACL}, 2019.

\bibitem{dosovitskiy2020image}
Alexey Dosovitskiy, Lucas Beyer, Alexander Kolesnikov, Dirk Weissenborn,
  Xiaohua Zhai, Thomas Unterthiner, Mostafa Dehghani, Matthias Minderer, Georg
  Heigold, Sylvain Gelly, Jakob Uszkoreit, and Neil Houlsby.
\newblock An image is worth 16x16 words: Transformers for image recognition at
  scale.
\newblock In {\em ICLR}, 2021.

\bibitem{dosovitskiy2015flownet}
Alexey Dosovitskiy, Philipp Fischer, Eddy Ilg, Philip Hausser, Caner Hazirbas,
  Vladimir Golkov, Patrick Van Der~Smagt, Daniel Cremers, and Thomas Brox.
\newblock Flownet: Learning optical flow with convolutional networks.
\newblock In {\em Proceedings of the IEEE International Conference on Computer
  Vision}, pages 2758--2766, 2015.

\bibitem{feichtenhofer_iccv_2019}
Christoph Feichtenhofer, Haoqi Fan, Jitendra Malik, and Kaiming He.
\newblock Slowfast networks for video recognition.
\newblock In {\em ICCV}, 2019.

\bibitem{gabeur2020multi}
Valentin Gabeur, Chen Sun, Karteek Alahari, and Cordelia Schmid.
\newblock Multi-modal transformer for video retrieval.
\newblock In {\em European Conference on Computer Vision}, pages 214--229.
  Springer, 2020.

\bibitem{girdhar2022omnivore}
Rohit Girdhar, Mannat Singh, Nikhila Ravi, Laurens van~der Maaten, Armand
  Joulin, and Ishan Misra.
\newblock Omnivore: A single model for many visual modalities.
\newblock {\em arXiv preprint arXiv:2201.08377}, 2022.

\bibitem{hershey2017cnn}
Shawn Hershey, Sourish Chaudhuri, Daniel~PW Ellis, Jort~F Gemmeke, Aren Jansen,
  R~Channing Moore, Manoj Plakal, Devin Platt, Rif~A Saurous, Bryan Seybold,
  et~al.
\newblock Cnn architectures for large-scale audio classification.
\newblock In {\em IEEE International Conference on Acoustics, Speech and Signal
  Processing (ICASSP)}, pages 131--135, 2017.

\bibitem{huang_stochasticdepth_eccv_2016}
Gao Huang, Yu Sun, Zhuang Liu, Daniel Sedra, and Kilian Weinberger.
\newblock Deep networks with stochastic depth.
\newblock In {\em ECCV}, 2016.

\bibitem{kay_arxiv_2017}
Will Kay, Joao Carreira, Karen Simonyan, Brian Zhang, Chloe Hillier, Sudheendra
  Vijayanarasimhan, Fabio Viola, Tim Green, Trevor Back, Paul Natsev, et~al.
\newblock The kinetics human action video dataset.
\newblock In {\em arXiv preprint arXiv:1705.06950}, 2017.

\bibitem{kazakos2021little}
Evangelos Kazakos, Jaesung Huh, Arsha Nagrani, Andrew Zisserman, and Dima
  Damen.
\newblock With a little help from my temporal context: Multimodal egocentric
  action recognition.
\newblock {\em BMVC}, 2021.

\bibitem{kazakos2019epic}
Evangelos Kazakos, Arsha Nagrani, Andrew Zisserman, and Dima Damen.
\newblock Epic-fusion: Audio-visual temporal binding for egocentric action
  recognition.
\newblock In {\em Proceedings of the IEEE/CVF International Conference on
  Computer Vision}, pages 5492--5501, 2019.

\bibitem{kondratyuk2021movinets}
Dan Kondratyuk, Liangzhe Yuan, Yandong Li, Li Zhang, Mingxing Tan, Matthew
  Brown, and Boqing Gong.
\newblock Movinets: Mobile video networks for efficient video recognition.
\newblock In {\em CVPR}, 2021.

\bibitem{monfort_pami_2019}
Mathew Monfort, Alex Andonian, Bolei Zhou, Kandan Ramakrishnan, Sarah~Adel
  Bargal, Tom Yan, Lisa Brown, Quanfu Fan, Dan Gutfreund, Carl Vondrick, et~al.
\newblock Moments in time dataset: one million videos for event understanding.
\newblock In {\em PAMI}, 2019.

\bibitem{nagrani2021attention}
Arsha Nagrani, Shan Yang, Anurag Arnab, Aren Jansen, Cordelia Schmid, and Chen
  Sun.
\newblock Attention bottlenecks for multimodal fusion.
\newblock In {\em NeurIPS}, 2021.

\bibitem{park2019specaugment}
Daniel~S Park, William Chan, Yu Zhang, Chung-Cheng Chiu, Barret Zoph, Ekin~D
  Cubuk, and Quoc~V Le.
\newblock Specaugment: A simple data augmentation method for automatic speech
  recognition.
\newblock {\em arXiv preprint arXiv:1904.08779}, 2019.

\bibitem{simonyan_neurips_2014}
Karen Simonyan and Andrew Zisserman.
\newblock Two-stream convolutional networks for action recognition in videos.
\newblock In {\em NeurIPS}, 2014.

\bibitem{steiner2021augreg}
Andreas Steiner, Alexander Kolesnikov, , Xiaohua Zhai, Ross Wightman, Jakob
  Uszkoreit, and Lucas Beyer.
\newblock How to train your vit? {D}ata, augmentation, and regularization in
  vision transformers.
\newblock In {\em arXiv preprint arXiv:2106.10270}, 2021.

\bibitem{stroud2020learning}
Jonathan~C. Stroud, Zhichao Lu, Chen Sun, Jia Deng, Rahul Sukthankar, Cordelia
  Schmid, and David~A. Ross.
\newblock Learning video representations from textual web supervision.
\newblock In {\em arXiv 2007.14937}, 2020.

\bibitem{szegedy_cvpr_2016}
Christian Szegedy, Vincent Vanhoucke, Sergey Ioffe, Jon Shlens, and Zbigniew
  Wojna.
\newblock Rethinking the inception architecture for computer vision.
\newblock In {\em CVPR}, 2016.

\bibitem{touvron2021training}
Hugo Touvron, Matthieu Cord, Matthijs Douze, Francisco Massa, Alexandre
  Sablayrolles, and Herv{\'e} J{\'e}gou.
\newblock Training data-efficient image transformers \& distillation through
  attention.
\newblock In {\em ICML}, 2021.

\bibitem{vaswani2017attention}
Ashish Vaswani, Noam Shazeer, Niki Parmar, Jakob Uszkoreit, Llion Jones,
  Aidan~N Gomez, {\L}ukasz Kaiser, and Illia Polosukhin.
\newblock Attention is all you need.
\newblock In {\em NeurIPS}, 2017.

\bibitem{wu2022memvit}
Chao-Yuan Wu, Yanghao Li, Karttikeya Mangalam, Haoqi Fan, Bo Xiong, Jitendra
  Malik, and Christoph Feichtenhofer.
\newblock Memvit: Memory-augmented multiscale vision transformer for efficient
  long-term video recognition.
\newblock {\em arXiv preprint arXiv:2201.08383}, 2022.

\bibitem{yan2022multiview}
Shen Yan, Xuehan Xiong, Anurag Arnab, Zhichao Lu, Mi Zhang, Chen Sun, and
  Cordelia Schmid.
\newblock Multiview transformers for video recognition.
\newblock In {\em CVPR}, 2022.

\end{thebibliography}
}

\end{document}